\pdfoutput=1

\documentclass[11pt]{article}

\usepackage[]{ACL2023}

\usepackage{times}
\usepackage{latexsym}

\usepackage[T1]{fontenc}

\usepackage[utf8]{inputenc}

\usepackage{microtype}

\usepackage{inconsolata}
\usepackage{makecell}
\usepackage{multirow}
\usepackage{amsfonts}
\usepackage{amsmath}
\usepackage{booktabs}
\usepackage{mathrsfs}
\usepackage{graphicx}

\title{Exploiting Abstract Meaning Representation for \\Open-Domain Question Answering}

\vspace{5mm}
\author{
Cunxiang Wang\textsuperscript{$\spadesuit$$\clubsuit$\thanks{\ \ \  Done during the internship at Amazon AWS AI.}}, 
Zhikun Xu\textsuperscript{$\heartsuit$}, 
Qipeng Guo\textsuperscript{$\diamondsuit$}, 
Xiangkun Hu\textsuperscript{$\diamondsuit$},
\AND 
Xuefeng Bai\textsuperscript{$\clubsuit$}, 
Zheng Zhang\textsuperscript{$\diamondsuit$} and Yue Zhang\textsuperscript{$\clubsuit$\thanks{\ \ \ The correponding author.}}
\vspace{3mm}
\\
\textsuperscript{$\spadesuit$}Zhejiang University, China\\
\textsuperscript{$\clubsuit$}School of Engineering, Westlake University, China\\
\textsuperscript{$\heartsuit$}Fudan University, China;
\textsuperscript{$\diamondsuit$}Amazon AWS AI\\
  {\{wangcunxiang, zhangyue\}@westlake.edu.cn}
  }

\begin{document}
\maketitle
\begin{abstract}
The Open-Domain Question Answering (ODQA) task involves retrieving and subsequently generating answers from fine-grained relevant passages within a database. Current systems leverage Pretrained Language Models (PLMs) to model the relationship between questions and passages. However, the diversity in surface form expressions can hinder the model's ability to capture accurate correlations, especially within complex contexts. Therefore, we utilize Abstract Meaning Representation (AMR) graphs to assist the model in understanding complex semantic information. We introduce a method known as Graph-as-Token (GST) to incorporate AMRs into PLMs. Results from Natural Questions (NQ) and TriviaQA (TQ) demonstrate that our GST method can significantly improve performance, resulting in up to 2.44/3.17 Exact Match score improvements on NQ/TQ respectively. Furthermore, our method enhances robustness and outperforms alternative Graph Neural Network (GNN) methods for integrating AMRs. To the best of our knowledge, we are the first to employ semantic graphs in ODQA. \footnote{\ \ We release our code and data at https://github.com/wangcunxiang/Graph-aS-Tokens }
\end{abstract}

\section{Introduction}

\begin{figure*}[t]
  \center{
  \vspace{-14mm}
  \includegraphics
  [width=16cm]  
  {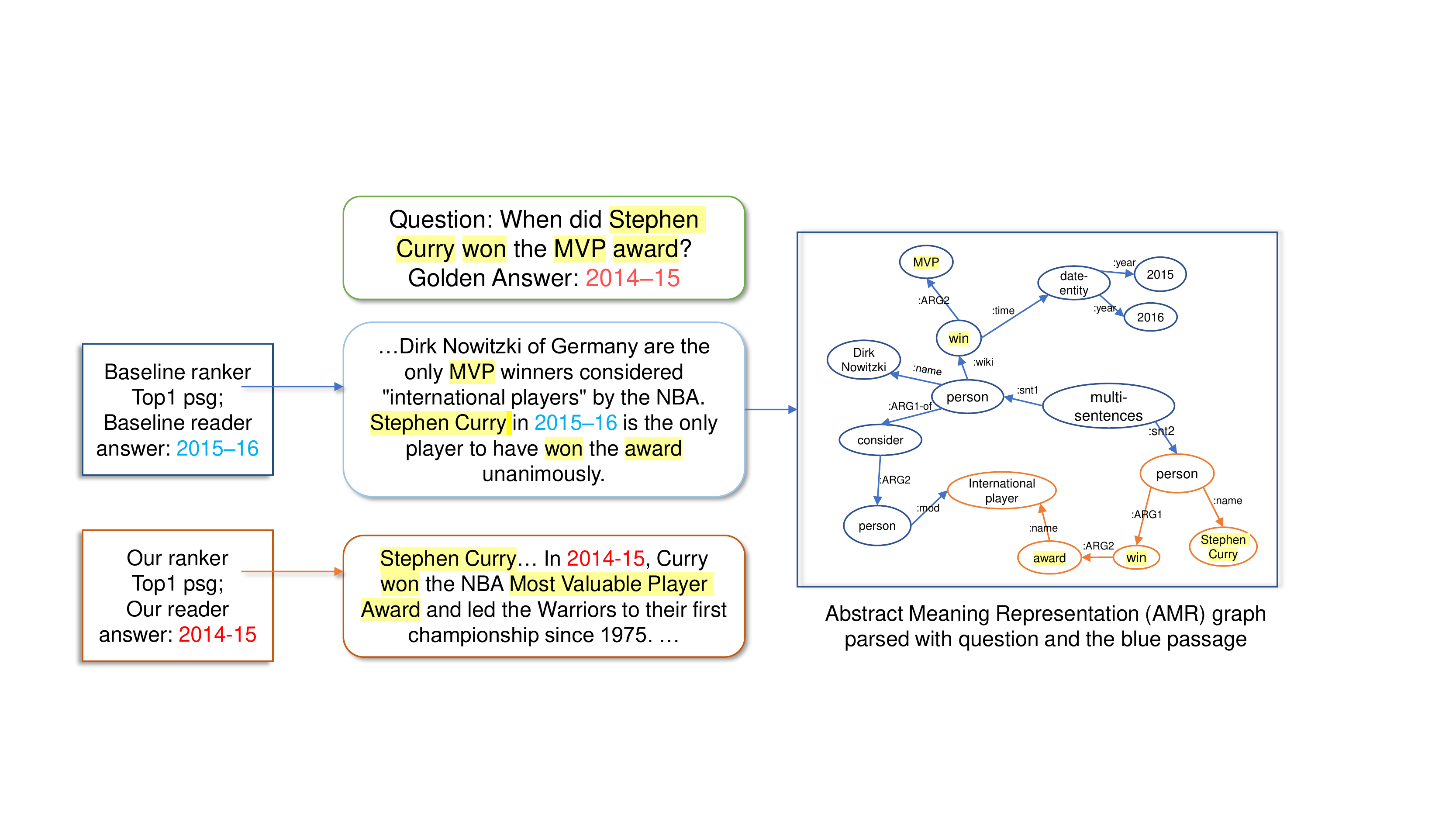}}
  \vspace{-20mm}
  \caption{
  An example from our experiments. The top-middle square contains the question and the gold standard answer. The middle section shows a confusing passage with an incorrect answer generated by the baseline model and ranked first by the baseline reranker. The bottom-middle section presents a passage with the gold standard answer, which is ranked within the top ten by our reranker but not by the baseline. Important information is highlighted.}
  \label{main_example}
  \vspace{-2mm}
\end{figure*}

Question Answering (QA) is a significant task in Natural Language Processing (NLP) \citep{SQuAD}. Open-domain QA (ODQA) \citep{DrQA}, particularly, requires models to output a singular answer in response to a given question using a set of passages that can total in the millions. ODQA presents two technical challenges: the first is \textit{retrieving} \citep{DPR} and \textit{reranking} \citep{R2D2} relevant passages from the dataset, and the second is generating an answer for the question using the selected passages. In this work, we focus on the \textit{reranking} and \textit{reading} processes, which necessitate fine-grained interaction between the question and passages.

Existing work attempts to address these challenges using Pretrained Language Models (PLMs) \citep{Re2G}. However, the diverse surface form expressions often make it challenging for the model to capture accurate correlations, especially when the context is lengthy and complex. We present an example from our experiments in Figure~\ref{main_example}. In response to the question, the reranker incorrectly ranks a confusing passage first, and the reader generates the answer \textit{``2015–16''}. The error arises from the PLMs' inability to effectively handle the complex semantic structure. Despite \textit{``MVP''}, \textit{``Stephen Curry''} and \textit{``won the award''} appearing together, they are not semantically related. In contrast, in the AMR graph, it is clear that \textit{``Stephen Curry''} wins over \textit{``international players''}, not the \textit{``MVP''}, which helps the model avoid the mistake. The baseline model may fail to associate "Most Valuable Player" in the passage with "MVP" in the question, which may be why the baseline does not rank it in the Top10. To address this issue, we adopt structured semantics (i.e., Abstract Meaning Representation \citep{AMR} graphs shown on the right of Figure~\ref{main_example}) to enhance Open-Domain QA.

While previous work has integrated graphs into neural models for NLP tasks, adding additional neural architectures to PLMs can be non-trivial, as training a graph network without compromising the original architecture of PLMs can be challenging \citep{ribeiro}. Converting AMR graphs directly into text sequences and appending them can be natural, but leads to excessively long sequences, exceeding the maximum processing length of the transformer. To integrate AMR into PLMs without altering the transformer architecture and at a manageable cost, we treat nodes and edges of AMR \underline{G}raphs a\underline{S} \underline{T}okens (GST) in PLMs. This is achieved by projecting the embeddings of each node/edge, which consist of multiple tokens, into a single token embedding and appending them to the textual sequence embeddings. This allows for integration into PLMs without altering the main model architecture. This method does not need to integrate a Graph Neural Network into the transformer architecture of PLMs, which is commonly used in integrating graph information into PLMs \citet{KG-FiD, GRAPE}. The GST method is inspired by \citet{Kim2022PureTA} in the graph learning domain, who uses token embeddings to represent nodes and edges for the transformer architecture in graph learning tasks. However, their method is not tailored for NLP tasks, does not consider the textual sequence embeddings, and only handles a certain types of nodes/edges, whereas we address unlimited types of nodes/edges consisting of various tokens.

Specifically, we select BART and FiD as baselines for the reranking and reading tasks, respectively. To integrate AMR information, we initially embed each question-passage pair into text embeddings. Next, we parse the pair into a single AMR graph using AMRBART \citep{AMRBART}. We then employ the GST method to embed the graph nodes and graph edges into graph token embeddings and concatenate them with the text embeddings. Lastly, we feed the concatenated text-graph embeddings as the input embeddings to a BART-based \citep{BART} reranker to rerank or a FiD-based \citep{FiD} reader to generate answers.

We validate the effectiveness of our GST approach using two datasets -- Natural Question \citep{NQ} and TriviaQA \citep{TQ}. Results indicate that AMR enhances the models' ability to understand complex semantics and improves robustness. BART-GST-reranker and FiD-GST outperform BART-reranker and FiD on the reranking and reading tasks, respectively, achieving up to 5.9 in Top5 scores, 3.4 in Top10 score improvements, and a 2.44 increase in Exact Match on NQ. When the test questions are paraphrased, models equipped with GST prove more robust than the baselines. Additionally, GST outperforms alternative GNN methods, such as Graph-transformer and Relational Graph Convolution Network (RGCN) \citep{RGCN}, for integrating AMR.

To the best of our knowledge, we are the first to incorporate semantic graphs into ODQA, thereby achieving better results than the baselines.

\section{Related Work}

\paragraph{Open-domain QA.}
Open-Domain Question Answering (ODQA) \citep{DrQA} aims to answer one factual question given a large-scale text database, such as Wikipedia.
It consists of two steps. The first is \textit{dense passage retrieval} \citep{DPR} , which retrieves a certain number of passages that match the question. 
In this process, a \textit{reranking} step can be used to filter out the most matching passages \citep{R2D2, Re2G}. 
The second is \textit{reading}, which finds answer by reading most matching passages \citep{FiD, RAG}.
We focus on the reranking and reading, and integrate AMR into those models.

\paragraph{Abstract Meaning Representation (AMR)} \citep{AMR} is a formalism for representing the semantics of a text as a rooted, directed graph. In this graph, where nodes represent basic semantic units such as entities and predicates, and edges represent the relationships between them.
Compared with free-form natural language, AMR graphs are more semantically stable as sentences with same semantics but different expressions can be expressed as the same AMR graph \citep{amr4dial, DocAMR}. 
In addition, AMR graphs are believed to have more structure semantic information than pure text \citep{DocAMR}. 

Previous work has implemented AMR graphs into neural network models. For example, \citep{amr4dial} adopts Graph-transformer \citep{graph_transformer} to integrate AMRs into the transformer architecture for the dialogue understanding and generation. AMR-DA \citep{AMRDA} uses AMRs as an data augmentation approach which first feeds the text into AMRs and regenerates the text from the AMRs. \citet{SARA} uses AMR graphs with rich semantic information to redesign the pre-training tasks which results in improvement on downstream dialogue understanding tasks. 
However, none of them is used for Open-domain QA or applied with the GST technique. which does not need to implement extra architectures in the PLMs, avoiding the incompatibility of different model architectures.

\paragraph{Integrating Structures into PLMs for ODQA}

Some work also tries to integrate structure information into PLMs for ODQA. For example, GRAPE \citep{GRAPE}  insert a Relation-aware Graph Neural Network into the T5 encoders of FiD to encode knowledge graphs to enhance the output embeddings of encoders;  KG-FiD \citep{KG-FiD} uses the knowledge graph to link different but correlated passages, reranks them before and during the reading, and only feeds the output embeddings of most correlated passages into the decoder.
However, existing work concentrates on the knowledge graph as the source of structure information and no previous work has considered AMRs for ODQA.

\begin{figure*}[t]
  \center{
  \includegraphics
  [width=14cm]  
  {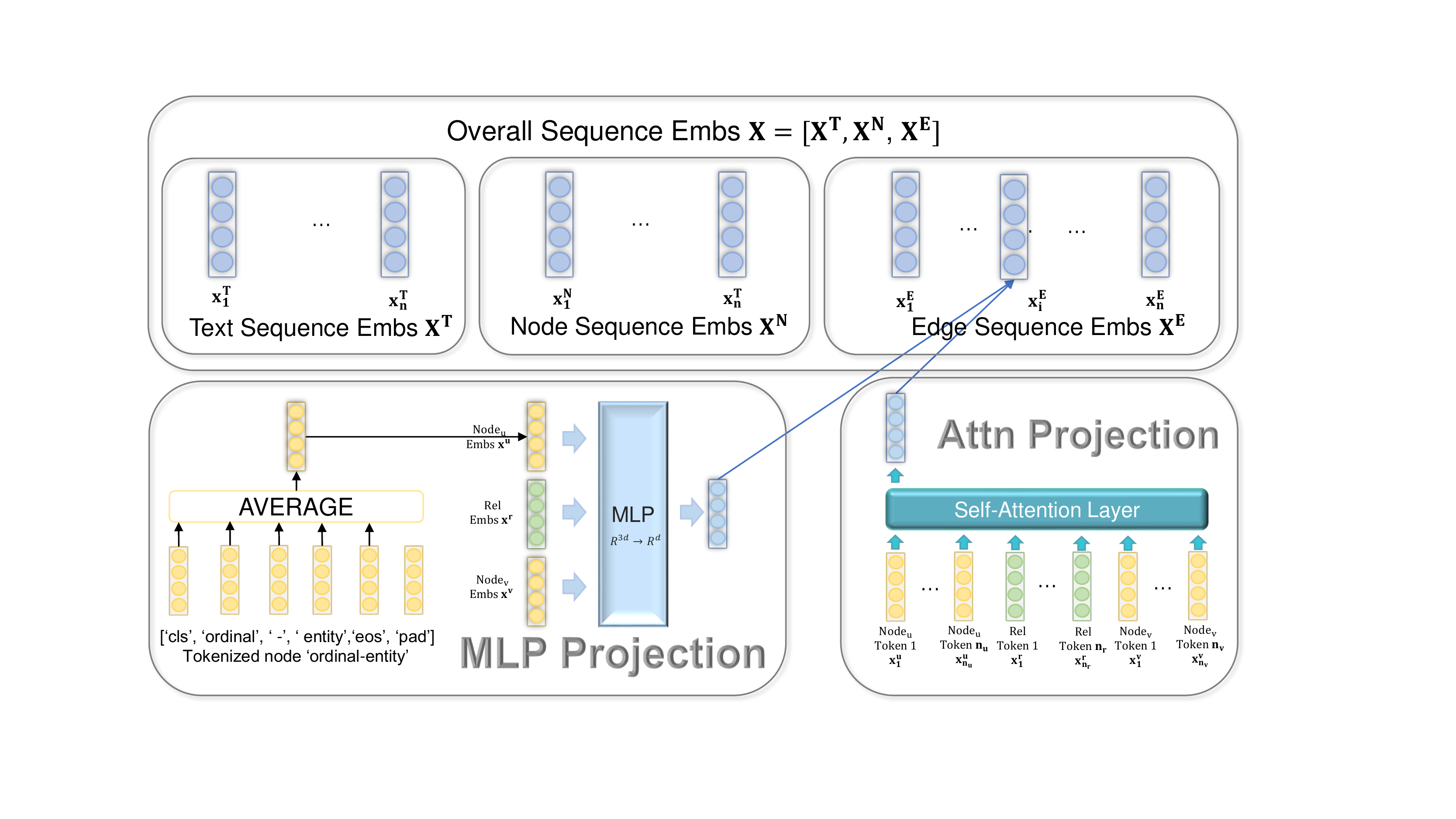}}
  \vspace{-6mm}
  \caption{The structure of our Graph-aS-Token method. The input consists of the text and the AMR graph of one passage; The output is a united embedding.}
  \label{model}
  \vspace{-2mm}
\end{figure*}

\paragraph{LLMs in Open-Domain Question Answering (ODQA)}
Research has been conducted that utilizes pre-trained language models (PLMs) to directly answer open-domain questions without retrieval \citep{yu2023generate,wang-etal-2021-generative,ye2021studying,rosset2021pretrain}. The results, however, have traditionally not been as effective as those achieved by the combined application of DPR and FiD. It was not until the emergence of ChatGPT that direct answer generation via internal parameters appeared to be a promising approach.

In a study conducted by \citet{wang2023evaluating}, the performances of Large Language Models (LLMs), such as ChatGPT (versions 3.5 and 4), GPT-3.5, and Bing Chat, were manually evaluated and compared with that of DPR+FiD across NQ and TQ test sets. The findings demonstrated that FiD surpassed ChatGPT-3.5 and GPT-3.5 on the NQ test set and outperformed GPT-3.5 on the TQ test set, affirming the relevance and effectiveness of the DPR+FiD approach even in the era of LLMs.

\section{Method}
We introduce the Retrieval and Reading of Open-Domain QA and their baselines in Section~\ref{baseline},  AMR graph generation in Section~\ref{amr_generation} and our method Graph-aS-Token (GST) in Section~\ref{GST}. 

\subsection{Baseline}
\label{baseline}

\paragraph{Retrieval.}
The retrieval model aims to retrieve $N_1$ passages from $M$ reference passages  ($N_1 << M$) given the question $q$.
Only fast algorithms, such as BM25 and DPR \citep{DPR}, can be used to retrieve from the large-scale database, and complex but accurate PLMs cannot be directly adopted. So, retrieval algorithm is often not very accurate. 
One commonly used method is applying a reranking process to fine-grain the retrieval results, and we can use PLMs to encode the correlations, which is usually more accurate.
Formally, reranking requires model to sort out the most correlated $N_2$ passages with $q$ from $N_1$ passages ($N_2 < N_1$). For each passage $p $ in the retrieved passage $P_{N_1}$, we concatenate the $q$ $p$ together and embed them into text sequence embeddings  $X_{qp} \in \mathbb{R}^{L \times H} $, where $L$ is the max token length of the question and passage pair and $H$ is the dimension.

We use a pretrained language model to encode each $\mathbf{X_{qp}}$ and a classification head to calculate a correlation score between $q$ and $p$:
\begin{equation}
\begin{split}
s_{qp} =  PLM (\mathbf{X_{qp}})
\end{split}
\end{equation}
where $PLM$ denotes the pretrained language model and the commonly used  Multi-Layer Perceptron (MLP) is used as as the classification head. 

We use the cross entropy as the loss function,
\begin{equation}
    \begin{aligned}
    \mathcal{L} 
    &= \frac{1}{N_q} \sum_{q} [ \frac{1}{N_{pos} + N_{neg}} \sum_{p} l_{qp}] \\
    &= \frac{1}{N_q * (N_{pos} + N_{neg})} \sum_{q} \sum_{p} -\\
    & [(y_{qp} * log(s_{qp}) 
    + (1-y_{qp}) * log(1-s_{qp}))],
    \end{aligned}
\end{equation}
where $N_{pos}$ and $N_{neg}$ are the numbers of positive and negative passages for training one question, respectively. To identify positive/negative label of each passage to the question, we follow \citet{DPR},  checking whether at least one answer appears in the passage. 

We choose the $N_2$  passages which have reranked among Top-$N_2$ for the reading process.

\paragraph{Reading.}
The reader needs to generate an answer $a$ given the question $q$ and $N_2$ passages. In this work, we choose the Fusion-in-Decoder (FiD) model \citep{FiD} as the baseline reader model.
The FiD model uses $N_2$ separate T5 encoders \citep{t5} to encode $N_2$ passages and concatenate the encoder hidden states to feed in one T5 decoder to generate answer.

Similar to reranking, we embed the question $q$ and each passage $p$ to text sequence embeddings $\mathbf{X_{qp}} \in \mathbb{R}^{L \times d_H} $, where $L$ is the max token length of the question and passage pair and $d_H$ is the dimension.
Next, we feed the embeddings in the FiD model to generate the answer
\begin{equation}
    a = FiD ([\mathbf{X_{qp_1},\dots, X_{qp_i},  X_{qp_{N_2}}}] )
\end{equation}
where $a$ is a text sequence.

\subsection{AMR}
\label{amr_generation}
We concatenate each question $q$ and passage $p$, parse the result sequence into an AMR graph $G_{qp} = \{V, E\}$, where $V, E$ are nodes and edges, respectively. Each edge is equipped with types, so $e = \{(u, r, v)\}$ where $u, r, v$ represent the head node, relation and the tail node, respectively.

\subsection{Graph aS Token (GST)}
\label{GST}

As shown in Figure~\ref{model}, we project each node $n$ or edge $e$ in one AMR graph $G$ into node embedding $\mathbf{x^n}$ or edge embedding $\mathbf{x^e}$. We adopt two types of methods to project each node and edge embeddings to one token embedding, which are MLP projection and Attention projection.
After the projection, we append the node embeddings $\mathbf{X^N = [x^n_1, \dots, x^n_{n_n}]}$ and edge embeddings $\mathbf{X^E = [x^e_1, \dots, x^e_{n_e}]}$ to the corresponding text sequence embeddings $\mathbf{X^T = [x^t_1, \dots, x^t_{n_t}]}$. So, the result sequence embedding is in the following notation:
\begin{equation}
    \mathbf{X = [X^T, X^N, X^E]}
\end{equation}

\paragraph{Initialization}
We explain how we initialize embeddings of nodes and edges here.

As each node $n$ and relation $r$  contain plural tokens (example of node `ordinal-entity' is shown the left and bottom of Figure~\ref{model}), $n = [t1,.., t_n]$ and $r = [t1, \dots, t_r]$, and each edge $e$ contains two nodes and one relation, we have $e = [[t1,.., t_{u}], [t1, \dots, t_r], [t1,.., t_{v}]]$. 

For edges and nodes, we first embed their internal tokens into token embedding. 

For edges, we have
\begin{equation}
\begin{aligned}
\mathbf{x^{e1}} =&\mathbf{[[x^{u}_1}, \dots, \mathbf{x^{u}_{n_u}]},\\
&\mathbf{[x^{r}_1}, \dots, \mathbf{x^{r}_{n_r}]}, \\
&\mathbf{[x^{v}_1}, \dots, \mathbf{x^{v}_{n_v}]]}
\end{aligned}
\end{equation}

For nodes, we have
\begin{equation}
\mathbf{x^{n1} = [x^{n}_1, \dots, x^{n}_{n}]}
\end{equation}

\paragraph{MLP Projection}

The process is illustrated in the MLP Projection part of Figure~\ref{model}.
As each AMR node can have more than one tokens, we first average its token embeddings. For example, for a head node $u$, $\mathbf{x^{u}} = AVE([\mathbf{x^{u}_1}, \dots, \mathbf{x^{u}_{n_u}}]) \in \mathbb{R}^{d_H}$. The same is done for the relation.

Then, we concatenate the two node embeddings and one relation embedding together as the edge embedding, 
\begin{equation}
\begin{aligned}
\mathbf{x^{e2}}=[\mathbf{x^{u}, x^{r}, x^{v}}] \in \mathbb{R}^{3 d_H}
\end{aligned}
\end{equation}

Next, we use a $\mathbb{R}^{3d_H \times d_H}$ MLP layer to project the  $\mathbf{x^{e2}} \in \mathbb{R}^{d_H}$ into $\mathbf{x^{e}} \in \mathbb{R}^{d_H}$, and the final edge embedding
\begin{equation}
\begin{aligned}
\mathbf{x^{e}}
&=MLP(\mathbf{x^{e2}}) \\
&=MLP([\mathbf{x^{u}, x^{r}, x^{v}}])
\end{aligned}
\end{equation}

Similarly, we average the node tokens embeddings first $\mathbf{x^{n1}} = AVE([\mathbf{x^{n}_1, \dots, x^{n}_{n}}])$. To reuse the MLP layer, we copy the node embedding two times and concatenate, so, $\mathbf{x^{n2}} =[\mathbf{x^{n1}, x^{n1}, x^{n1}}] \in \mathbb{R}^{3 d_H}$. Last, We adopt an MLP layer to obtain final node embedding
\begin{equation}
\begin{aligned}
\mathbf{x^{n}}=MLP(\mathbf{x^{n2}}) \in \mathbb{R}^{d_H}
\end{aligned}
\end{equation}

We have also tried to assign separate MLP layers to nodes and edges, but preliminary experiments show that it does not improve the results.

\paragraph{Attention Projection}
We use one-layer self-attention to project nodes and edges into embeddings, which is shown in the Attn Projection part in Figure~\ref{model}.
The edge embedding is calculated
\begin{equation}
\begin{aligned}
\mathbf{x^{e}} &=Att_{E}([\mathbf{x^{u}_1, \dots, x^{u}_{n_u}},\\
&\mathbf{x^{r}_1, \dots, x^{r}_{n_r}, x^{v}_1, \dots, x^{v}_{n_v}}])
\end{aligned}
\end{equation}

Similarly, the node embedding is calculated
\begin{equation}
\begin{aligned}
\mathbf{x^{n}} &=Att_{N}([\mathbf{x^{n}_1, \dots, x^{n}_{n}}),
\end{aligned}
\end{equation}
where $Att_{E}$ and $Att_{N}$ both denote one self-attention layer for edges and nodes, respectively. We take the first token (additional token) embedding from the self-attention output as the final embedding.

We only modify the input embeddings from $\mathbf{X = X^T}$ to $\mathbf{X = [X^T, X^N, X^E]}$. The rest details of models, such as the transformer architecture and the training paradigm, are kept the same with the baselines.
Our model can directly use the PLMs to encode AMR graphs, without incompatibility between GNN's parameters and PLMs' parameters. 

\section{Experiments}

\begin{table*}[t]
  \centering
  \small
  \setlength{\tabcolsep}{1mm}
  \begin{tabular}{c|c|c|c|c|c|c}
  \toprule
   \multirow{3}{*}{Reranker + Reader $\backslash$ Dataset}
  & \multicolumn{3}{c|}{\makecell[c]{\textbf{Natural Questions}}}
  & \multicolumn{3}{c}{\makecell[c]{\textbf{TriviaQA}}}
  \\
  \cline{2-7}
  & \multicolumn{2}{c|}{\makecell[c]{Reranking}} & Reading
  & \multicolumn{2}{c|}{\makecell[c]{Reranking}} & Reading
  \\
  \cline{2-7}
  & 
  \makecell[c]{Top5}& \makecell[c]{Top10} & \makecell[c]{EM} &
  \makecell[c]{Top5}& \makecell[c]{Top10} & \makecell[c]{EM} \\
  \midrule
  \makecell[c]{w/o reranker + FiD-reader} & \multirow{3}{*}{73.7/74.6}  & \multirow{3}{*}{79.5/80.3}   & 49.47/50.66 &  \multirow{3}{*}{78.0/78.1}  &  \multirow{3}{*}{81.5/81.8} &  69.02/69.50  \\
  \makecell[c]{w/o reranker + FiD-GST-A} & &  & 50.12/51.11  & &  &  70.17/70.39  \\
  \makecell[c]{w/o reranker + FiD-GST-M} &  & &  50.06/50.97 & &  &  69.98/70.10  \\
  \midrule
  \makecell[c]{BART-reranker + FiD-reader} &  \multirow{3}{*}{78.7/78.6} & \multirow{3}{*}{83.0/83.3}   &  50.33/51.33 &  \multirow{3}{*}{83.2/83.2} &  \multirow{3}{*}{85.2/85.1}  & 71.16/71.33 \\
  \makecell[c]{BART-reranker + FiD-GST-A}  &  & & 50.80/52.38&  &   & 71.93/72.05 \\
  \makecell[c]{BART-reranker + FiD-GST-M}&   & &  50.76/52.24  & &   & 72.12/72.24 \\
  \midrule
  \makecell[c]{BART-GST-A + FiD-reader} &\multirow{2}{*}{79.3/79.3}  & \multirow{2}{*}{83.3/83.3}& 50.68/52.18& \multirow{2}{*}{\textbf{83.5/83.3}}  & \multirow{2}{*}{\textbf{85.3/85.3}} &  71.54/71.71 \\
  \makecell[c]{BART-GST-A + FiD-GST-A} &  &  & 51.05/52.80   & &  & \textbf{72.63/72.67} \\
  \midrule
  \makecell[c]{BART-GST-M + FiD-reader} & \multirow{2}{*}{\textbf{79.6/80.0}}   &  \multirow{2}{*}{83.3/\textbf{83.7}}  &   51.11/52.13 & \multirow{2}{*}{83.1/82.9} & \multirow{2}{*}{85.0/85.1} & 71.47/71.62\\
  \makecell[c]{BART-GST-M + FiD-GST-M} &    & & \textbf{51.40/53.10} &  && 72.58/72.61\\
  \bottomrule
  \end{tabular}
  \caption{Reranking and reading results on the dev/test set of NQ and TQ. In each cell, the left is on the dev while the right is on the test. For the BART/FiD with GST-M/A in the first column, they are equipped AMR graphs with the GST method, -M indicates the MLP projection while -A is the attention projection. 
  }
  \label{main_table}
\end{table*}

\begin{table}[t]
  \centering
  \small
  \setlength{\tabcolsep}{0.4mm}
  \begin{tabular}{c|c|c|c|c}
  \toprule
   \multirow{2}{*}{Rearnker $\backslash$ Dataset}
  & \multicolumn{2}{c|}{\makecell[c]{\textbf{Natural Questions}}}
  & \multicolumn{2}{c}{\makecell[c]{\textbf{TriviaQA}}}
  \\
  \cline{2-5}
  & 
  \makecell[c]{MRR}& \makecell[c]{MH@10} & \makecell[c]{MRR}& \makecell[c]{MH@10}\\
\midrule
  \makecell[c]{w/o reranker }& 20.2/18.0 & 37.9/34.6  & 12.1/12.3 & 25.5/25.9 \\
  \midrule
  \makecell[c]{BART-reranker}&25.7/23.3  & 49.3/45.8   &  16.9/17.0 & 37.7/38.0   \\
  \midrule
  \makecell[c]{BART-GST-A}& 28.1/24.7 & 52.7/48.2   &  \textbf{17.7/17.8} & \textbf{39.3/39.9}  \\
  \midrule
  \makecell[c]{BART-GST-M}  &\textbf{28.4/25.0} & \textbf{53.2/48.7}  & 17.5/17.6 &  39.1/39.5  \\
  \bottomrule
  \end{tabular}
  \caption{Overall reranking results on NQ and TQ. In each cell, the left is dev and the right is test. 
  }
  \label{mrr}
\end{table}

\subsection{Data}
We choose two representative Open-Domain QA datasets, namely Natural Questions (NQ) and TriviaQA (TQ), for experiments. Data details are in presented in Appendix Table~\ref{appendix:dataset}.

Since retrieval results have a large impact on the performance of downstream reranking and reading, we follow \citet{FiD} and \citep{KG-FiD} to fix retrieval results for each experiment to make the reranking and reading results comparable for different models. In particular, we use the DPR model initialized with parameters in \citet{izacard2020distilling} 
\footnote{\url{https://dl.fbaipublicfiles.com/FiD/pretrained_models/nq_retriever.tar.gz} \\ \url{https://dl.fbaipublicfiles.com/FiD/pretrained_models/tqa_retriever.tar.gz}} to retrieve 100 passages for each question. Then we rerank them into 10 passages, which means $N_1 = 100, N_2 = 10$.

We generate the amr graphs using  AMRBART \citep{AMRBART} (the AMRBART-large- finetuned-AMR3.0-AMRParsing checkpoint) \footnote{https://huggingface.co/xfbai/AMRBART-large-finetuned-AMR3.0-AMRParsing}.

\subsection{Models Details}
We choose the BART model as the reranker baseline and the  FiD model (implemented on T5 model\citep{t5}) as the reader baseline, and adopt the GST method on them. 
For each model in this work, we use its Large checkpoint, such as BART-large and FiD-large, for reranking and reading, respectively.
In the reranking process, we evaluate the model using the dev set per epoch, and use Top10 as the pivot metric to select the best-performed checkpoint for the test.
For the reading, we evaluate the model per 10000 steps, and use Exact Match as the pivot metric.
For training rerankers, we set number of positive passages as 1 and number of negative passages as 7.
We run experiments on 2 Tesla A100 80G GPUs.

\subsection{Metric}
Following \citet{Re2G} and \citet{FiD}, we use Top-N to indicate the reranking performance and Exact Match for the reading performance. 

However, TopN is unsuitable for indicating the overall reranking performance for all positive passages, so we also adopt two metrics, namely Mean Reciprocal Rank (MRR) and Mean Hits@10 (MHits@10). 
The MRR score is the Mean Reciprocal Rank of all positive passages. Higher scores indicate that the  positive passages are ranked higher overall. 
The MHits@10 indicates the percentage of positive passages are ranked in Top10. Higher scores indicate that more  positive passages are ranked in Top10. 
Their formulations  are in Appendix Section~\ref{appendix:metircs}.
Note that, only when the retrieved data is exactly the same, the MRR and MHits@10 metrics are comparable. 

\subsection{Preliminary Experiments}

We present the reranking performance of four baseline PLMs, including BERT \citep{BERT}, RoBERTa \citep{RoBERTa}, ELECTRA \citep{electra} and BART \citep{BART} on the NQ and TQ in Appendix Table~\ref{appendix:pre-exp}.
BART outperforms other three models in every metric on both NQ and TQ. So, we choose it as the reranker baseline and apply our Graph-aS-Token method to it in following reranking experiments.

\subsection{Main Results}

The Main results are presented in Table~\ref{main_table}. 
Our method can effectively boost the performance on both reranking and reading.

\paragraph{Reading.}
As shown in the reading columns of  Table~\ref{main_table}, our method can boost the FiD performance, no matter whether there is reranker and whether the reranker is with AMR or not.
Without reranking, FiD-GST-A achieves 51.11/70.39 EM on NQ/TQ test , which are 0.45/0.89 EM higher over the baseline FiD;
With reranking, `BART-GST-M + FiD-GST-M ' achieves 53.10/72.61 EM on NQ/TQ test, 1.77/1.27 EM better than `BART-reranker + FiD'. 
With the same reranker, FiD-GST is better than the baseline FiD, for example, `BART-reranker + FiD-GST-A' achieves  52.38/72.05 on NQ/TQ test, which is 1.05/0.72 higher than  
the 51.33/71.33 of `BART-reranker + FiD'.

Overall, our GST models have achieved up to 2.44 EM (53.10 vs 50.66) on NQ test and 3.17 (72.67 vs 69.50) on TQ test.

\paragraph{Reranking}
Shown in the reranking columns of Table~\ref{main_table}, BART-GST-M can achieve 80.0/83.7 scores in Top5/Top10, which  improve 5.4/3.4 on NQ-test compared to DPR and 1.4/0.4 compared to BART-reranker.
BART-GST-M achieves 79.3/83.3 scores in Top5/Top10, which outperform DPR by 4.7/3.0 on NQ-test, showing that our GST method is effective.

We present results of the MRR and MHits@10 metrics in Table~\ref{mrr}. Our GST method can help positive passages rank higher in Top10. In NQ, BART-GST-M has 7.0/14.1 advantages on MRR/MHits@10 over DPR while 1.7/2.9 advantages over BART-reranker; In TQ, BART-GST-A has 5.5/14.0 advantages on MRR/MHits@10 over DPR and 0.8/1.9 advantages on MRR, MHits@10 over BART-reranker.

The overall reranking results can also explain the reason why even when the Top10 results are similar and readers are the same, the reranked passages by BART-GST can lead to better reading performance. For example, in NQ test, the reading performance of `BART-GST-M + FiD'  is 0.80 better than `BART-reranker + FiD'.

\begin{table}[t]
  \centering
  \small
\setlength{\tabcolsep}{2mm}
  \begin{tabular}{c|c|c|c}
  \toprule
  & 
  \makecell[c]{Orig Test}& \makecell[c]{New Test} & Drop \\
  \midrule
  BART-reranker  & \makecell{78.6/83.3\\23.3/45.8}  & \makecell{76.2/81.8\\21.5/43.6}  & \makecell{-2.6/-1.5\\-1.8/-2.2} \\
  \midrule
  BART-GST-A & \makecell{79.3/83.3\\24.7/48.2 } & 
  \makecell{77.4/82.0\\23.2/46.1} & 
  \makecell{\textbf{-1.9}/-1.3\\
  \textbf{-1.4}/\textbf{-2.1}} \\
  \midrule
  BART-GST-M & 
  \makecell{80.0/83.7\\25.0/48.7} &
  \makecell{78.0/82.4\\23.4/46.3} &
  \makecell{-2.0/-1.3\\-1.6/-2.4}\\
  \bottomrule
  \multicolumn{4}{c}{\makecell{A: Robustness of rerankers. Each cell contains \\Top5/Top10/MRR/MHits@10 as the metrics. }}\\
  \toprule
  & 
 \makecell[c]{Orig Test}& \makecell[c]{New Test} & Drop\\
  \midrule
  FiD-reader &  50.66 & 46.76 &-3.90 \\
  \midrule
  FiD-GST-A & 51.11 & 47.84&-3.27 \\
  \midrule
  FiD-GST-M & 50.97 & 47.76 & \textbf{-3.21} \\
  \bottomrule
  \multicolumn{4}{c}{\makecell{B: Robustness of readers. Exact Match as the Metric. \\To avoid the influence of different reranking results, \\we use the same DPR results to train and eval.}}\\
  \end{tabular}
\caption{Robustness on rerankers and readers. We conduct experiments on NQ.  \textit{Orig Test} is the original test questions while \textit{New Test} means the paraphased test questions. \textit{Drop} is the difference from the original test to the paraphrased test, the smaller absolute number indicates better robustness.}
\vspace{-2mm}
  \label{robusteness}
\end{table}

\subsection{Analysis}

\paragraph{Robustness.}
To evaluate the robustness of the baseline and our models, we paraphrase the test questions of NQ and TQ, evaluate 
paraphrased test questions and the original ones with the same model checkpoint.
We use a widely-used paraphraser, namely \textit{Parrot Paraphraser} \citep{prithivida2021parrot} to paraphrase test questions. The results are shown in Table~\ref{robusteness}. 

The performance drops in reranking and reading of our GST models are lower than the baseline model, despite that our models have better performance. For reranking, the drop of our BART-GST-A is -1.9/-1.3/-1.4/-2.1 for Top5/Top10/MRR/MHits@10, which is lower than the baseline's -2.6/-1.5/-1.8/-2.2. For reading, the -3.21 EM drop of FiD-GST-M is also smaller than the -3.90 of baseline FiD. It shows that our GST method can not only improve performance but also improve robustness, which can prove that adding structural information can help models avoid the erroneous influence of sentence transformation.

\paragraph{Comparison with FiD-100.}
\begin{table}[t]
  \centering
  \small
\setlength{\tabcolsep}{0.3mm}
  \begin{tabular}{c|c|c|c|c}
  \toprule
  & NQ dev & NQ test & TQ dev & TQ test
  \\
  \midrule
  FiD-10  & 49.47 & 50.66 & 69.02 & 69.50 \\
  \midrule
  FiD-100 & \textbf{51.60} & 52.88 & 71.61 & 71.88 \\
  \midrule
  \makecell{FiD-10\\w/ BART-reranker} & 50.33 & 51.33 & 71.16 & 71.33 \\
  \midrule
  \makecell{FiD-GST-A-10\\w/ BART-GST-A reranker}& 51.03 & 52.80 & \textbf{72.63} & \textbf{72.67} \\
  \midrule
  \makecell{FiD-GST-M-10\\w/ BART-GST-M reranker} & 51.30 & \textbf{53.10} & 72.58 & 72.61 \\
  \bottomrule
  \end{tabular}
\caption{Reading experiments of with and without reranking. The first two row are trained/evaluated with DPR data while the rest are with reranked data. } 
\vspace{-2mm}
  \label{FiD-100}
\end{table}

We also compare the reranking+reading paradigm with the directly-reading paradigm. For the latter, the FiD reader is directly trained and evaluated on 100 retrieved passages without reranking.
The results are shown in Table~\ref{FiD-100}. 

Without our GST method, the reranking+reading paradigm (FiD-10 w/ BART reranker) is worse than FiD-100 without reranking, which is 71.33 to 71.78 on the test. However, with our GST method, the reranking+reading paradigm outperforms FiD-100. For example, FiD-GST-M-10 w/ BART-GST-M reranker has better performance on NQ test than FiD-100, which is 53.10 vs 52.88, and FiD-GST-A-10 w/ BART-GST-A reranker vs FiD-100 on TQ test is 72.67 vs 71.78.

To our knowledge, we are the first make FiD-10 beat FiD-100.

\paragraph{Influence of AMR Quality.}
\begin{table}[t]
  \centering
  \small
\setlength{\tabcolsep}{0.5mm}
  \begin{tabular}{c|c|c|c|c}
  \toprule
  & Top5 & Top10 & MRR & MH@10
  \\
  \midrule
  BART-reranker  & 78.7/78.6 &83.0/83.3& 25.7/23.3&49.3/45.8 \\
  \midrule
  \makecell{BART-GST-M\\(superior AMRs)}  &79.6/80.0 & 83.3/83.7 & 28.4/25.0&53.2/48.7 \\
  \midrule
  \makecell{BART-GST-M\\(inferior AMRs)}  & 79.5/79.3 & 83.5/83.1 & 28.4/24.7 & 52.9/47.8  \\
  \bottomrule
  \multicolumn{5}{c}{\makecell{In reranking.}}\\
  \toprule
  & \multicolumn{4}{c}{Exact Match} \\
  \midrule
  FiD-reader & \multicolumn{4}{c}{48.47/50.66} \\
  \midrule
  \makecell{FiD-GST-A\\(superior AMRs)}  &  \multicolumn{4}{c}{50.12/51.11} \\
  \midrule
  \makecell{FiD-GST-A\\(inferior AMRs)}  &  \multicolumn{4}{c}{49.95/50.83} \\
  \bottomrule
  \multicolumn{5}{c}{\makecell{In reading.}}
  \end{tabular}
\caption{Influence of superior AMR graphs which generated by a larger model, and inferior AMR graphs which generated by a smaller model. } 
\vspace{-2mm}
  \label{amr-quality}
\end{table}
We explore how AMR graphs quality influence the performance of our models in this section, 
by using the AMRBART-base-finetuned-AMR3.0-AMRParsing, \footnote{https://huggingface.co/xfbai/AMRBART-base-finetuned-AMR3.0-AMRParsing} which is a smaller version.
We compare the reranking performance of BART-GST with either superior or inferior graphs on NQ and TQ. We use the each kind of graphs to train its own reranking models. The results are shown in Table~\ref{amr-quality}.

Our models still work with inferior AMR graphs but the performance is not good as the superior ones in  both reranking and reading.
This indicates that when the quality of AMR graphs is higher, the GST models can potentially achieve better performance.

\paragraph{Ablation to Nodes/Edges}
\begin{table}[t]
  \centering
  \small
\setlength{\tabcolsep}{0.5mm}
  \begin{tabular}{c|c|c|c|c}
  \toprule
  & Top5 & Top10 & MRR & MH@10
  \\
  \midrule
  BART-reranker  & 78.7/78.6 &83.0/83.3& 25.7/23.3&49.3/45.8 \\
  \midrule
  BART-GST-M & 79.6/80.0 & 83.3/83.7 & 28.4/25.0&53.2/48.7\\
  \midrule
  \makecell{BART-GST-M\\only nodes} & 78.5/78.9 & 82.9/83.1 & 27.6/24.2 & 51.8/47.3 \\
  \midrule
  \makecell{BART-GST-M\\only edges} & 78.6/79.3 & 83.0/83.3 & 27.9/24.7 & 52.4/47.4 \\
  \bottomrule
  \end{tabular}
\caption{Ablation to nodes and edges to our GST methods on NQ. We choose BART-GST-M because it better performs on NQ.} 
\vspace{-2mm}
  \label{only-nodes-edges}
\end{table}

We ablate nodes and edges in our models to explore whether nodes or edges contribute more to the results. We conduct reranking experiments on NQ. The results are shown in Table~\ref{only-nodes-edges}.
As can be seen, nodes are edges are both useful for the GST method, where `BART-GST-M (only nodes)' and `BART-GST-M (only edges)' both outperform the baseline BART-reranker in MRR/MHits@10 on NQ test, which are 24.2/48.7 vs 24.7/47.4 vs 23.3/45.8, respectively.
However, `BART-GST-M (only edges)' are better than `BART-GST-M (only nodes)' in four metrics on NQ,  partly due to the fact that edges also contain nodes information.

\paragraph{Case Study}  
\begin{figure}[t]
  \center{
  \includegraphics
  [width=19cm]  
  {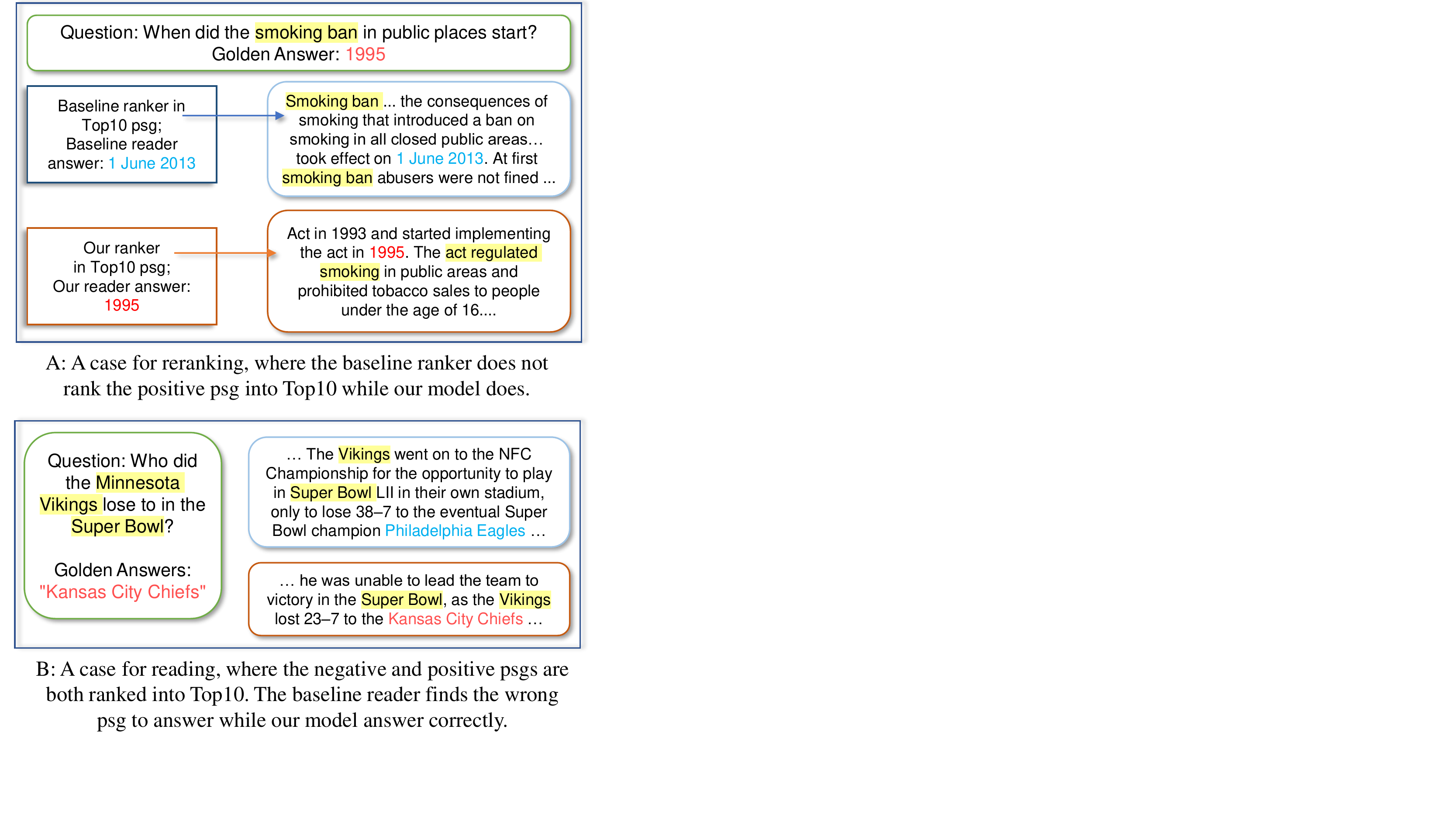}}
  \vspace{-15mm}
  \caption{Two cases from our experiments for reranking and reading, respectively. We highlight important information over questions and passages.}
  \label{case_study}
  \vspace{-4mm}
\end{figure}

We present two cases from our  in Figure~\ref{case_study}. 
In the upper one, for the negative passage, the baseline may consider \textit{``a ban on smoking in all closed public areas''} same as \textit{``the smoking ban in public places''}, which are actually different; For the positive passage, the baseline may not take \textit{``act regulated smoking in public area''} as \textit{``the smoking ban in public places''} while our model does.

In the lower one, the baseline reader ignores the competition is \textit{`` for the opportunity to play in Super Bow''} rather than \textit{``in the Super Bowl''} , and because the number of similar passages with \textit{``Philadelphia Eagle''} are more than the positive passage's, the baseline reader finds the incorrect passage which leads to the incorrect answer. In contrast, our model focuses on the only positive passage and answers the question correctly.

\subsection{Alternative Graph Methods}
\begin{table}[t]
  \centering
  \small
\setlength{\tabcolsep}{0.5mm}
  \begin{tabular}{c|c|c|c|c}
  \toprule
  & Top5 & Top10 & MRR & MH@10
  \\
  \midrule
  BART-reranker  & 78.7/78.6 &83.0/83.3& 25.7/23.3&49.3/45.8 \\
  \midrule
  BART-GST-M & 79.6/80.0 & 83.3/83.7 & 28.4/25.0&53.2/48.7\\
  \midrule
  RGCN-Stacking & 78.6/78.2 & 82.3/83.0 & 26.1/23.1 & 49.5/46.0 \\
  \bottomrule
  \end{tabular}
\caption{Comparison between the baseline, GST and RGCN-Stacking in reranking on NQ.} 
\vspace{-2mm}
  \label{GNN_results}
\end{table}

We have also tried several methods to integrate AMRs into PLMs, but their performance is worse than our Graph-aS-Token method. Here we take two representative examples, which are Relational Graph Convolution Network (RGCN) \citep{RGCN} for the reranker and Graph-transformer \citep{graph_transformer} for FiD. All those methods require alignments between text tokens and graph nodes, for which only some nodes can be successfully aligned. 

\paragraph{\textbf{Stacking RGCN above Transformer}}

The model architecture consists of a transformer encoder and a RGCN model where RGCN is stacked on top of the transformer. After the vanilla forward by transformer encoder, AMR graphs abstracted from queries and passages in advance are constructed with node embeddings initialized from transformer output. Then they are fed into the RGCN model and the final output of the [CLS] node is used for scoring.

For the text embeddings of one question-passage pair, its encoder hidden states
$$
    \mathbf{H} = Encoder (X_{qp})
$$

For one node $n$, its initial embedding
$$
    \mathbf{h^{0}} = MeanPooling ( \mathbf{H_{start:end}})
$$
where $start$ and $end$ are the start and end positions of the text span aligned with the node.

The update of node embedding for each layer $l$ is
$$
    \mathbf{h^{l+1}_{i}} = \sigma (W^{l}_{0} \mathbf{h^{l}_{i}} + \sum_{r \in R} \sum_{j\in N^{r}_i} \frac{1}{c_{i,r}} W^{l}_r \mathbf{h^{l}_{i}})
$$
$$
    c_{i,r} = \| N^r_i \|
$$
where $R$ is the set of edge types, $N^r_i$ stands for the group of nodes which connect with node $i$ in relation $r$.

so the correlation score of $q$ and $p$:
$$
s_{qp} = ClsHead(h^{L}_{[CLS]})
$$

The results are presented in Table~\ref{GNN_results}, which is clear that the RGCN-stacking method is inferior to the GST method. Some metrics, including Top5, Top10 and MRR, of RGCN-stacking are worse than the baseline, meaning the RGCN method is not feasible for integrating AMRs into PLMs though it looks like reasonable and practical.

\paragraph{\textbf{Graph-transformer}}
We apply the graph-transformer architecture to FiD model for reading.
We follow the graph-transformer architecture in \citet{amr4dial}, whose main idea is using AMR information to modify the self-attention scores between text tokens.
However, we find stucking challenging for PLMs because the new-initialized graph architectures are not compatible  with architectures of PLMs, lead to non-convergence during training.  
Despite that, tricks such as incrementally training and separate tuning can lead to convergence, results are still below the baseline model, let alone GST.

\paragraph{\textbf{Flattening AMR Graphs}}
We have also tried to directly flatten AMR graphs into text sequences, but the result sequences are always beyond the maximum processing length (1024) of the transformer. So, we have to cut off some nodes and edges to fit in the transformer, but the results show that it does not work well and has only a very sight improvement  while the computational cost is tens times over the baseline.

\section{Conclusion}
In this study, we successfully incorporated Abstract Meaning Representation (AMR) into Open-Domain Question Answering (ODQA) by innovatively employing a Graph-aS-Token (GST) method to assimilate AMRs with pretrained language models. The reranking and reading experiments conducted on the Natural Questions and TriviaQA datasets have demonstrated that our novel approach can notably enhance the performance and resilience of Pretrained Language Models (PLMs) within the realm of ODQA.

\section*{Acknowledgement}
This publication has emanated from research conducted with the financial support of the Pioneer and ``Leading Goose" R\&D Program of Zhejiang under Grant Number 2022SDXHDX0003.

\section*{Limitations}
Our Graph-aS-Token (GST) method can increase the time and GPU memory cost, we set an quantitative analysis in Appendix Section \ref{sec_cost}. 
We train the models with only one random seed.
We do not conduct a large number of hyper-parameter tuning experiments, but use a fixed set of hyper-parameters to make the baseline and our models comparable.

\section*{Ethics Statement}
No consideration.

\bibliography{anthology,acl2023}
\bibliographystyle{acl_natbib}

\appendix



\section{Experimental Details}

\subsection{Pre-experiment}

\begin{table}[t]
  \centering
  \small
\setlength{\tabcolsep}{0.5mm}
  \begin{tabular}{c|c|c|c|c}
  \toprule
  & Top5 & Top10 & MRR & MH@10
  \\
  \midrule
  w/o reranker  & 73.7/74.6 & 79.5/80.3 & 20.2/18.0 & 37.9/34.6 \\
  \midrule
  BERT & 76.5/75.7 & 81.5/81.4 & 23.7/20.9 & 45.5/41.5 \\
  \midrule
  RoBERTa & 77.1/76.6 & 82.3/82.3 & 24.7/21.5 & 47.7/43.3 \\
  \midrule
  ELECTRA & 77.3/77.8 & 82.4/82.5 & 25.1/22.5 & 47.9/43.9 \\
  \midrule
  BART & \textbf{78.7/78.6} & \textbf{83.0/83.3} & \textbf{25.7/23.3} & \textbf{49.3/45.8} \\
  \bottomrule
  \multicolumn{5}{c}{A: On the Natural Questions dataset.}\\
  \toprule
   & Top5 & Top10 & MRR & MH@10 \\
  \midrule
  w/o reranker  & 78.0/78.1 & 81.5/81.8 & 12.1/12.3 & 25.5/25.9 \\
  \midrule
  BERT & 82.0/82.3 & 84.5/84.7 & 16.0/16.2 & 35.6/35.9 \\
  \midrule
  RoBERTa & 82.8/82.9 & 85.0/85.0 & 16.8/16.8 & 37.2/37.4 \\
  \midrule
  ELECTRA &82.4/82.6& 84.8/82.6& 16.3/16.4 & 36.2/36.4 \\
  \midrule
  BART & \textbf{83.2/83.1} & \textbf{85.2/85.1} & \textbf{16.9/17.0} & \textbf{37.7/38.0} \\
  \bottomrule
  \multicolumn{5}{c}{\makecell{B: On the TriviaQA dataset.}}\\
  \end{tabular}
\caption{Pre-experiments of four PLMs' reranking performance on NQ and TQ. In each cell, the left is on the dev while the right is on the test. Among four PLMs, BART performs best.}
\vspace{-2mm}
  \label{appendix:pre-exp}
\end{table}

\subsection{Details for Data}
\begin{table}[t]
  \centering
  \small
  \setlength{\tabcolsep}{1mm}
  \begin{tabular}{c|c|c|c}
  \toprule
   &
  \makecell[c]{\textbf{Train Set}} & 
  \makecell[c]{\textbf{Dev Set}} & 
  \makecell[c]{\textbf{Test Set}}  \\
  \midrule
  Natural Questions & 79168 & 8757 & 3610 \\
  \midrule
  TriviaQA & 78785 & 8837 & 11313\\
  \bottomrule
  \end{tabular}
  \caption{Details of each dataset. 
  }
  \vspace{-3mm}
  \label{appendix:dataset}
\end{table}

For each question and passage pair, we feed it in the generator in such a format ``Question: <question>. Title: <Passage Title>. Context: <Passage Context>''.
Additionally, we link the nodes, which are recognized as entities such as person name and date and have same surfaces, with the ``:same'' relation because it helps performance.
For nodes in one AMR graph, we remove their `-XX', where X is a 0-9 number.

\subsection{Hyper-parameters}

We set other model-related hyper-parameters in Table~\ref{appendix:Hyper-parameters}.

\begin{table}[t]
  \centering
  \small
  \setlength{\tabcolsep}{0.5mm}
  \begin{tabular}{c|c|c}
  \toprule
   &
  \makecell[c]{\textbf{Reranking}} & 
  \makecell[c]{\textbf{Reading}} \\
  \midrule
  Leaning Rate & 3e-5 & 1e-4  \\
  \midrule
  Training Epoch & 10 & 5 \\
  \midrule 
  Node MaxLength & 145 & 145 \\
  \midrule
  Edge MaxLength & 165 & 165 \\
  \midrule 
  Text Maxlength & 200 & 200 \\
  \midrule
  Eval Step/Epoch & 10k steps & 1 epoch \\
  \bottomrule
  \end{tabular}
  \caption{Hyper-parameters Setting 
  }
  \vspace{-2mm}
  \label{appendix:Hyper-parameters}
\end{table}

\subsection{Cost Increase}
\label{sec_cost}

\begin{table}[t]
  \centering
  \small
  \setlength{\tabcolsep}{0.5mm}
  \begin{tabular}{c|c|c}
  \toprule
   &
  \makecell[c]{Time cost} & 
  \makecell[c]{GPU Memory Cost} \\
  \midrule
  FiD & 1.00 & 1.00  \\
  \midrule
  FiD-GST-M & 1.29 & 1.11 \\
  \midrule 
  FiD-GST-M & 1.40 & 1.40 \\
  \bottomrule
  \end{tabular}
  \caption{The results of time and GPU memory cost comparing our GST method and the baseline. The experiment is inference on the NQ test set. We take the baseline FiD model cost as 1.00.
  }
  \vspace{-2mm}
  \label{appendix:cost}
\end{table}

We conduct an experiment of the increase of time and GPU memory cost on our GST compared with the baseline. For inference, while keeping other parameters as the same, the time costs of FiD-GST-M, FiD-GST-A are 1.29x and 1.40x, respectively, and the GPU memory costs are 1.11x and 1.40x, respectively, compared with FiD, as shown in Table \ref{appendix:cost}.

\subsection{Metrics}
\label{appendix:metircs}
$$
    MRR = \frac{1}{|Q|} \sum_{i \in Q} (({\sum_{j\in Pos}\frac{1}{t(j)}})\frac{1}{num_{Pos}(i)})
$$
where $Q$ is the evaluating dataset; $t(j)$ is the rank of passage $j$; $Pos$ is the set of positive passages.

$$
    MHits@10 = \frac{1}{|Q|} \sum_{i \in Q} ({\sum_{j\in pos,t(j)<=10} \frac{1}{num_{Pos}(i)}})
$$
where $Q$ is the evaluating dataset; $t(j)$ is the rank of passage $j$; $Pos$ is the set of positive passages.

\end{document}